\documentclass[letterpaper, 10 pt, conference]{ieeeconf}
\IEEEoverridecommandlockouts    
\usepackage{graphics}           
\usepackage{times}              
\usepackage{amsmath}            
\usepackage{amssymb}            
\usepackage{graphicx}
\usepackage{algorithm}
\usepackage[noend]{algpseudocode}
\usepackage{booktabs}
\usepackage{color}
\usepackage{listings}
\usepackage{subfiles}
\usepackage{hyperref}
\usepackage[nocompress]{cite} 
\definecolor{instructioncolor}{rgb}{.5,.5,.5}

\usepackage[font=small]{caption}

\def\secref#1{Sec.~\ref{#1}}
\def\figref#1{Fig.~\ref{#1}}
\def\tabref#1{Tab.~\ref{#1}}
\def\eqref#1{Eq.~(\ref{#1})}

\makeatletter
\usepackage{xspace}
\DeclareRobustCommand\onedot{\futurelet\@let@token\@onedot}
\def\@onedot{\ifx\@let@token.\else.\null\fi\xspace}

\makeatother

\usepackage{array}
\newcolumntype{L}[1]{>{\raggedright\let\newline\\\arraybackslash\hspace{0pt}}m{#1}}
\newcolumntype{C}[1]{>{\centering\let\newline\\\arraybackslash\hspace{0pt}}m{#1}}
\newcolumntype{R}[1]{>{\raggedleft\let\newline\\\arraybackslash\hspace{0pt}}m{#1}}

\renewcommand{\b}[1]{\mbox{\boldmath$#1$}}

\usepackage{multirow}
\usepackage[table]{xcolor}

\makeatletter
\def\blfootnote{\gdef\@thefnmark{}\@footnotetext}
\makeatother

\title{\LARGE \bf 3D Hierarchical Panoptic Segmentation \\ in Real Orchard Environments Across Different Sensors}

\author{Matteo Sodano \qquad Federico Magistri \qquad Elias Marks \qquad Fares Hosn \qquad Aibek Zurbayev \\ Rodrigo Marcuzzi \qquad Meher V. R. Malladi \qquad Jens Behley \qquad Cyrill Stachniss
  \thanks{All authors are with the Center for Robotics, University of Bonn, Germany. Cyrill Stachniss is additionally with the Department of Engineering Science at the University of Oxford, UK, and with the Lamarr Institute for Machine Learning and Artificial Intelligence, Germany.}%
  \thanks{This work has partially been funded 
  by the Deutsche Forschungsgemeinschaft (DFG, German Research Foundation) under Germany's Excellence Strategy, EXC-2070 -- 390732324 -- PhenoRob and under STA~1051/5-1 within the FOR 5351~(AID4Crops),
  by the European Union's Horizon Europe research and innovation programme under grant agreement No~101070405~(DigiForest),
  and by the BMBF in the project ``Robotics Institute Germany'', grant No.~16ME0999.
  }%
}

\begin{document}

\twocolumn[{%
\renewcommand\twocolumn[1][]{#1}%

\maketitle
\thispagestyle{empty}
\pagestyle{empty}

\begin{center}
    \centering
    \vspace{-1.5em}
    \captionsetup{type=figure}
    \includegraphics[width=.92\linewidth]{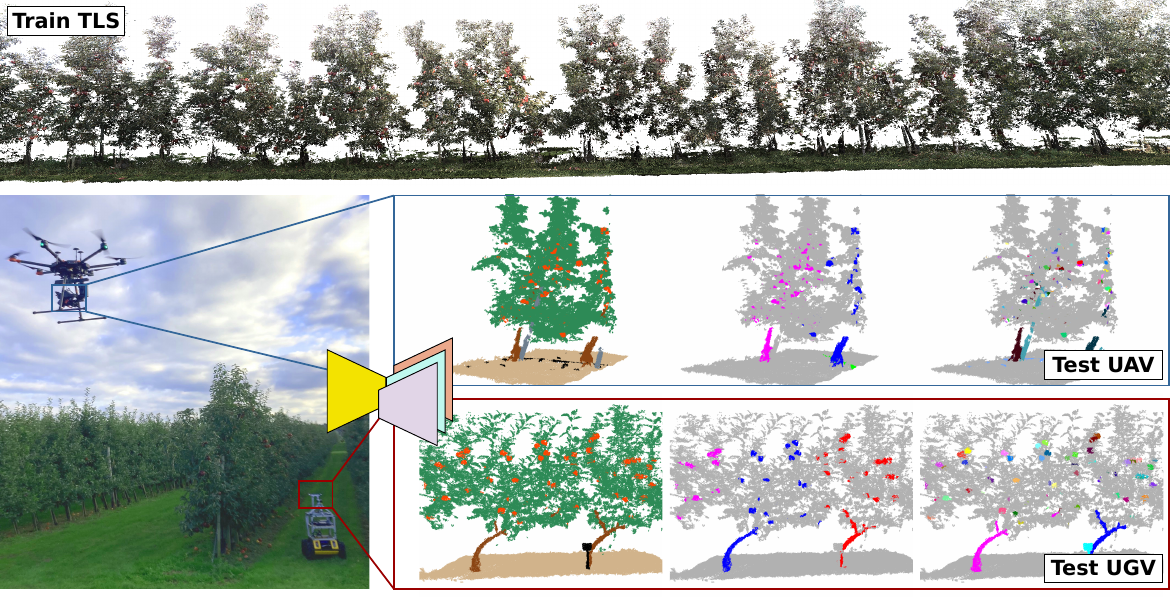}
    \captionof{figure}{View of the apple orchard (top row) recorded with a terrestrial laser scanner (TLS), and used for training our network for hierarchical panoptic segmentation. The test set is composed of different splits corresponding to different sensors and robots. The two examples shown in the image are a UAV equipped with a PhaseOne iXM-100 camera, and a UGV equipped with a RealSense d435i camera.}
    \label{fig:motivation}
\end{center}%
}]



\thispagestyle{empty}
\pagestyle{empty}

\begin{abstract}
  Crop yield estimation is a relevant problem in agriculture, because an accurate yield estimate can support farmers' decisions on harvesting or precision intervention. Robots can help to automate this process. To do so, they need to be able to perceive the surrounding environment to identify target objects such as trees and plants. In this paper, we introduce a novel approach to address the problem of hierarchical panoptic segmentation of apple orchards on 3D data from different sensors. Our approach is able to simultaneously provide semantic segmentation, instance segmentation of trunks and fruits, and instance segmentation of trees (a trunk with its fruits). This allows us to identify relevant information such as individual plants, fruits, and trunks, and capture the relationship among them, such as precisely estimate the number of fruits associated to each tree in an orchard.
  To efficiently evaluate our approach for hierarchical panoptic segmentation, we provide a dataset designed specifically for this task. Our dataset is recorded in Bonn, Germany, in a real apple orchard with a variety of sensors, spanning from a terrestrial laser scanner to a RGB-D camera mounted on different robots platforms. 
  The experiments show that our approach surpasses state-of-the-art approaches in 3D panoptic segmentation in the agricultural domain, while also providing full hierarchical panoptic segmentation. Our dataset is publicly available at \mbox{\url{https://www.ipb.uni-bonn.de/data/hops/}}. The open-source implementation of our approach is available at \url{https://github.com/PRBonn/hapt3D}.
  
\end{abstract}

\blfootnote{All authors are with the Center for Robotics, University of Bonn, Germany. Cyrill Stachniss is also with the Lamarr Institute for Machine Learning and Artificial Intelligence, Germany. 

This work has partially been funded by the German Federal Ministry of Education and Research (BMBF) in the project ``Robotics Institute Germany'', grant No.~\mbox{16ME0999}, by the Deutsche Forschungsgemeinschaft (DFG, German Research Foundation) under Germany's Excellence Strategy, EXC-2070 -- 390732324 -- PhenoRob and under \mbox{STA~1051/5-1} within the FOR 5351~(AID4Crops), and by the European Union's Horizon Europe research and innovation programme under grant agreement No~101070405~(DigiForest).}

\section{Introduction}
\label{sec:intro}

Crop yield estimation is an important task in agriculture. An accurate yield estimate can help farmers making management decisions to increase crop production and optimize key factors such as harvest time and fertilizer use~\cite{li2018plants, vanklompenburg2020cea, xu2019ei}, as well as enable precision interventions~\cite{blok2025cea}.
Robots can automatize or support many of these interventions, but to do so reliably and robustly, they need to understand their surroundings through the interpretation of sensor data.
Tasks such as segmenting, counting, and localizing fruits in orchards are crucial in horticulture, as they are required for replacing human fruit picking, which is an extremely labor-intensive process~\cite{calvin2010book}. 

Recently, many approaches targeted perception tasks in agriculture. Semantic segmentation~\cite{lottes2019jfr, milioto2018icra} and panoptic segmentation~\cite{marks2023ral, roggiolani2022icra, weyler2022ral} are especially common. This also motivated the publication of datasets for agricultural perception tasks~\cite{hani2020ral, kierdorf2022jfr, weyler2024pami}. However, 3D datasets of real agricultural environments are not common~\cite{dutagaci2020pm, schunck2021po}.


The main contributions of this paper are twofold. First, we propose an approach based on a convolutional neural network (CNN) for hierarchical panoptic segmentation, which is able to address multiple instance segmentation tasks at once while exploiting their underlying hierarchical relationship by means of a novel skip connections scheme. Second, we introduce a 3D point cloud dataset of real apple orchards, called HOPS (hierarchical orchard panoptic segmentation) recorded with various sensors, such as terrestrial laser scanner, RGB-D cameras, or RGB cameras with subsequent bundle adjustment. Additionally, we release high-quality annotations for hierarchical panoptic segmentation. We recorded data of the apple orchard at different growth stages across several crop rows in the span of two years, and labeled the resulting point clouds in order to obtain semantic segmentation, tree instance segmentation, as well as fruit and trunk instance segmentation. Specifically, in the tree instance segmentation each trunk has a label that is shared with all fruits belonging to it. Thus, in this instance-level task, fruits are not individually labeled. The fruit and trunk instance segmentation yields a different instance for each individual trunk and fruit. We show a portion of the dataset and an exemplary image of the approach we propose in~\figref{fig:motivation}.


Our experiments suggest that %
(i) our approach can jointly perform semantic, tree instance, and standard instance segmentation on real-world 3D data acquired with different sensors;
and
(ii) our skip connection scheme allows the CNN to exploit the underlying hierarchical relationship between the individual segmentation tasks, leading to better segmentation performance.
Our claims are backed up by the paper and our experimental evaluation. 

\section{Related Work}
\label{sec:related}


Semantic scene interpretation is often required for realizing robotic automation in agriculture, since it identifies task-relevant objects in the scene that enables applications such as monitoring~\cite{lottes2018iros, pan2023iros,smitt2024ral-pagn} and intervention~\cite{magistri2024ral,arad2020jfr,lehnert2020jfr}.

Panoptic segmentation~\cite{kirillov2019cvpr-ps} unifies semantic and instance segmentation provides semantics of background classes, so-called ``stuff'', and object-level instances of so-called ``things'' at the same time.
While often images are used for panoptic segmentation in the agricultural domain~\cite{weyler2022ral}, 3D point clouds generated by \mbox{RGB-D} cameras,  LiDARs, or photogrammetric structure-from-motion approaches received increasing research interest~\cite{zhu2024scidata} due to the capability to extract precise geometric information required for high-precision fruit grasping~\cite{magistri2024ral,lehnert2020jfr},  plant segmentation~\cite{montes2020iros}, and fine-grained trait estimation~\cite{marks2023ral}.
3D panoptic segmentation has been actively investigated in the domain of autonomous driving~\cite{sirohi2021tro,milioto2020iros,zhou2021cvpr-pplp}, where 
most approaches use dedicated branches for semantic and instance segmentation using features of a shared encoder. 
For plant phenotyping, we can exploit the hierarchical structure of plants that can be decomposed in individual parts~\cite{gueldenring2024ral}, such as plant, leaf and even more fine-grained leaf structure. This hierarchical structure has been exploited in the literature, leveraging skip connections between decoders for hierarchical bottom-up instance prediction tasks using images~\cite{roggiolani2022icra}, which is the foundation of our approach to 3D panoptic segmentation.

In contrast to prior approaches for 3D panoptic segmentation, we also explicitly consider the hierarchical structure of the prediction task by means of our novel skip connections scheme, which allows us to predict a hierarchical semantic scene interpretation consisting of plant instances as well as fruit instances that can be used for plant monitoring and yield estimation in the agricultural domain. While our approach is applied to orchard scenes, the hierarchical panoptic segmentation framework is general and can be applied to other domains, such as body part segmentation~\cite{lin2020tcsvt}.

While most agricultural datasets for semantic interpretation provide only RGB images~\cite{lu2020compag}, recently 3D datasets became available.
In particular, BUP20~\cite{smitt2024ral-pagn} provides an RGB-D dataset for instance segmentation of sweet peppers in a glasshouse~\cite{smitt2021icra}.
For 3D plant phenotyping in real agricultural fields, BonnBeetClouds~\cite{marks2024iros} provides annotated structure-from-motion point clouds consisting of plant and leaf instance of sugar beet plants.
Other datasets provide organ-level annotations of plants using point clouds acquired with high-precision LiDAR scanners~\cite{schunck2021po} or x-ray imaging~\cite{dutagaci2020pm} in the lab.
The Crops3D dataset~\cite{zhu2024scidata} provides organ-level annotations of different crop varieties for point cloud data acquired with a terrestrial laser scanner (TLS) in the field and single plant point clouds using structure-from-motion of RGB images or using structured light cameras.

In contrast to existing agricultural datasets for semantic scene interpretation, we provide a domain-specific agricultural dataset of  orchards recorded with different sensors that is annotated with plant and fruit instances.
We provide high-resolution point clouds recorded with a TLS, but also structure-from-motion point cloud data acquired with robotic platforms equipped with a high-resolution RGB camera or a consumer-grade RGB-D sensor commonly used in robotics.

\begin{figure*}
    \centering
    \includegraphics[width=.9\linewidth]{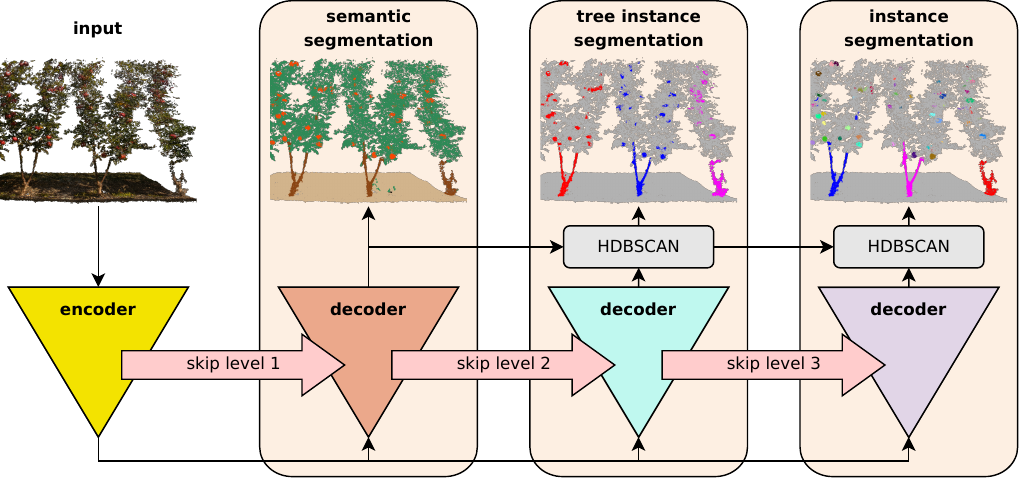}
    \caption{The encoder takes the colored point cloud as input, and the resulting features are processed by the decoders. We use hierarchical skip connections after each downsampling and upsampling block of the encoder and decoders. We use HDBSCAN to obtain instances.}
    \label{fig:architecture}
    \vspace{-1em}
\end{figure*}

\section{Our Approach \\ to Hierarchical Panoptic Segmentation}
\label{sec:main}
We propose an approach for hierarchical panoptic segmentation, i.e., the task of simultaneously performing semantic segmentation and multiple instance segmentation tasks with underlying hierarchical relationships. The network we use is an encoder-decoder architecture, and the decoders address semantic segmentation and two-levels of instance segmentation, as illustrated in~\figref{fig:architecture}. The first instance segmentation aims to identify ``tree'' instances, where a tree is defined by a trunk and all fruits belonging to it. The second instance segmentation looks for all individual instances (i.e., both trunks and fruits) in the point cloud.

\subsection{Architecture}
We use a MinkUNet-based neural network~\cite{choy2019cvpr} for our approach. This kind of neural networks are inspired by the UNet~\cite{ronneberger2015micc} design, where encoder and decoder are connected by skip connections. To process 3D data, MinkUnet networks use sparse 3D convolutions to process the input data. Specifically, we use the MinkUNet14A model, that allows us to have a lightweight network. In fact, MinkUNet14A is a modification of the standard MinkUNet14 model, and has fewer feature channels per layer, allowing faster computation. We keep the original structure of the network, and replicate the original decoder two additional times. This allows us to have three identical decoders addressing the three segmentation tasks we aim to tackle, which is a convenient solution for exploiting the underlying hierarchy among them.

\begin{figure*}
    \centering
    \includegraphics[width=.9\linewidth]{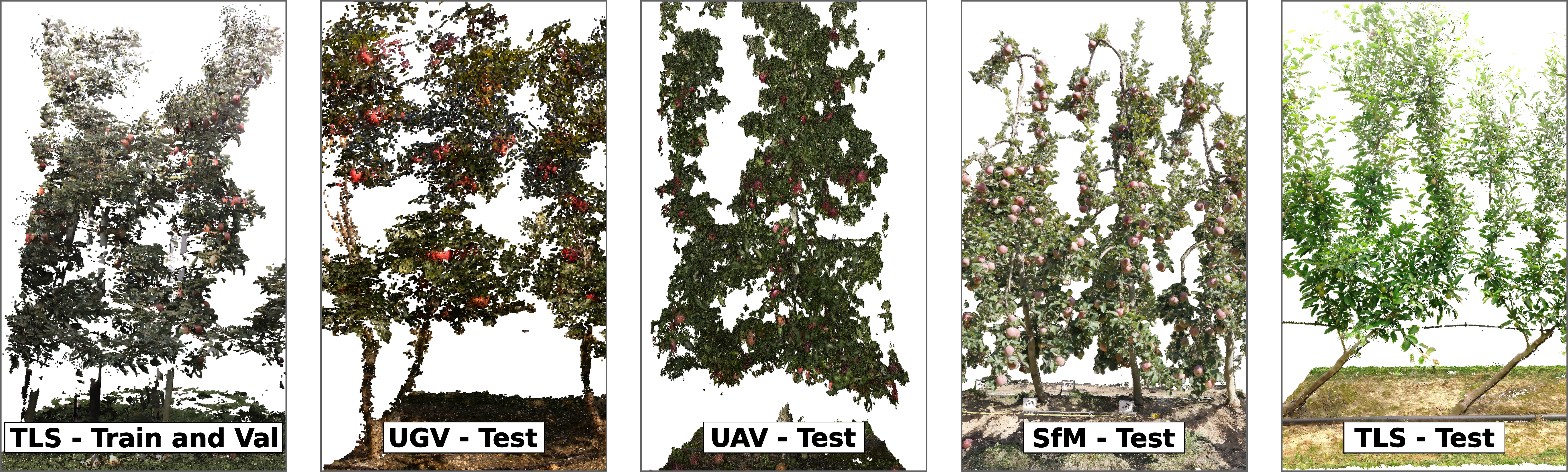}
    \caption{An example of the colored point clouds of HOPS. The point clouds from the train and validation sets are obtained using a TLS. We have four test sets recorded with different sensors. The one called ``SfM'' contains point clouds from the Fuji-SfM dataset~\cite{genemola2020compag,genemola2020db}.}
    \label{fig:data}
    \vspace{-1.5em}
\end{figure*}

The first decoder targets \textit{semantic segmentation}, and has a single output head with depth equal to the number of semantic classes that appear in the dataset. It is optimized using the standard weighted cross-entropy loss:
\begin{equation}
  \mathcal{L}_{\mathrm{sem}} = - \frac{1}{|\mathcal{S}|} \sum_{\b{p} \in \mathcal{S}} \omega_k \, \b{t}_{p}^\top \, \log(\sigma(\b{f}_p)),
\end{equation}
where $|\mathcal{S}|$ is the number of points in the point cloud, $\b{p}$ indicates the individual point, $\omega_k$ is a class-wise weight computed via the inverse frequency of each class in the dataset, where $k \in \{1, \, \dots, \, K\}$ indicates the semantic class, $\b{t}_p \in \mathbb{R}^{K}$ is the one-hot encoded ground truth annotation at point location $\b{p}$, $\sigma(\cdot)$ denotes the softmax operation, and $\b{f}_p$ denotes the pre-softmax feature predicted at point $\b{p}$.   

The second decoder targets \textit{tree instance segmentation}, i.e., the task of finding tree instances where a tree is defined as a trunk and all apples belonging to it. This decoder also has a single output head for offset prediction. Thus, each point predicts a 3D displacement towards a location in the 3D space that facilitates instance separation via clustering. We predict an offset $\b{o}_p \in \mathbb{R}^3$ for each point $\b{p}$, so that $\b{e}_p = \b{p} + \b{o}_p$ is the 3D location where the point is displaced. To obtain 3D points belonging to the same instance displaced to the same 3D location in space, and points belonging to different instances displaced to different locations, we adapt the Lov\'asz Hinge loss~\cite{neven2019cvpr} to 3D as:
\begin{equation}
  \mathcal{L}_{\mathrm{tree}} = \dfrac{1}{|\mathcal{C}|} \sum_{j=1}^{|\mathcal{C}|} \mathrm{Lov\acute{a}sz} (\b{F}_{\mathcal{C}_j}, \, \b{G}_{\mathcal{C}_j}),
  \label{eq:lovasz}
\end{equation}
where $\mathcal{C}$ is the set of object instances, $\b{G}_{\mathcal{C}_j} \in \{0, \, 1\}^{|\mathcal{S}|}$ denotes the binary ground truth mask of the $j$-th instance, and $\b{F}_{\mathcal{C}_j}$ is a soft-mask obtained by the offset prediction. The soft-mask $\b{F}_{\mathcal{C}_j}$ for instance $\mathcal{C}_j$ is obtained from the offset prediction: each point in the point cloud gets a score that depends on how far from the instance centroid $\b{c}_j$ its offset points to. The score is formalized as
\begin{equation}
  f_{\mathcal{C}_j} = \exp \bigg(- \frac{||\b{e}_p - \b{c}_j||^2}{2 \eta^2} \bigg),
\end{equation}
where $\b{e}_p$ indicates the 3D location pointed by the predicted offset at point $\b{p}$, and $\eta$ is a hyperparameter that defines an isotropic clustering region around the centroid.

Finally, the third decoder targets standard \textit{instance segmentation}, where we aim to individually segment each trunk and fruit in the scene. Similarly to the previous decoder, here also we have a single output head for 3D offset prediction, and optimize with the Lov\'asz Hinge loss $\mathcal{L}_{\mathrm{ins}}$, as in~\eqref{eq:lovasz}. 

Thus, the final loss function $\mathcal{L}$ is given by a weighted sum of the individual losses:
\begin{equation}
  \mathcal{L} = w_1 \, \mathcal{L}_{\mathrm{sem}} + w_2 \, \mathcal{L}_{\mathrm{tree}} + w_3 \, \mathcal{L}_{\mathrm{ins}}.
\end{equation}

\subsection{Skip Connections}
Skip connections are crucial to ensure feature reusability and address the gradient degradation problem of CNNs. They skip one or more layers, providing a direct gradient flow from the late stages of the decoder to the early stages of the encoder, which is usually hindered by the downsampling operation of the encoder.
We aim to extend our previous work on hierarchical panoptic segmentation on RGB images~\cite{roggiolani2022icra} to the 3D point cloud domain, and thus we propose to adopt a hierarchical skip connections scheme to address the underlying hierarchy of segmentation tasks. We propose to connect different decoders directly as shown in~\figref{fig:architecture}, instead of encoder and decoders only.

\textbf{Semantic segmentation}. For semantic segmentation, we keep the skip connection from the encoder to the decoder. The spatial information contained in the high-resolution feature maps of the encoder help the decoder for segmentation. 

\textbf{Tree instance segmentation}. For tree instance segmentation, we fuse the high-resolution maps coming from the encoder with the feature coming from the same level of the semantic decoder. In this way, the tree instance segmentation decoder will process an enriched feature map, containing also information about semantic segmentation. 

\textbf{Instance segmentation}. For instance segmentation, we fuse the high-resolution maps coming from the encoder with the feature coming from the same level of the tree instance segmentation decoder. Notice that, in this way, the skip features include semantic segmentation features as well, that come from the tree instance segmentation branch. 

To better illustrate the information flow in our hierarchical panoptic segmentation architecture, consider the process as assembling segmentation in layers. The encoder provides a basic high-resolution spatial guide. The semantic decoder adds labels to it. The tree instance segmentation decoder builds upon this enriched map, identifying individual trees as landmarks. Finally, the instance segmentation decoder refines this by adding finer distinctions, such as fruits and trunks, using all previously gathered information. At each stage, skip connections act like reference notes, ensuring that each level benefits from both low-level detail and high-level context, progressively enriching the representation.
This skip connection scheme allows us to exploit the hierarchy among tasks, enriching the features propagated to each decoder. In~\secref{sec:ablations}, we provide extensive experiments on different skip connection schemes to show that this design choice yields superior performance.

\subsection{Post-Processing}
As mentioned, the instance segmentation decoders predict an offset vector for each point. The goal is to have offsets of points belonging to the same instance indicating the same 3D location in space, and offsets of points belonging to different instances indicating different 3D locations. To obtain the final instance masks, we cluster the offsets with HDBSCAN~\cite{mcinnes2017joss}. Additionally, we use the semantic prediction to enforce consistency among instances in the third decoder, for example to avoid the case in which two points with two different semantic predictions end up in the same instance. This cannot be applied to the second decoder, as there instances are composed of one trunk and all apples attached to it, so a single instance actually includes multiple semantic classes.

\section{Our Dataset for \\ Hierarchical Panoptic Segmentation}
\label{sec:dataset}
Our dataset, called HOPS, is composed of point clouds collected with three different sensors at Campus Klein-Altendorf near Bonn, Germany. First, we collected point clouds using a TLS placed at multiple locations in the orchard row. Our training and validation set entirely contains point clouds collected with this sensor. Furthermore, we use a few TLS point clouds in the test set, collected in a different year and orchard location compared to the training and validation set. Second, we use point clouds obtained with a bundle adjustment procedure~\cite{triggs1999iccv} using as input images collected with a UAV equipped with a PhaseOne iXM-100 camera. We flew three missions on the same orchard with a camera angle of 45$^\circ$, 90$^\circ$, and 135$^\circ$ from the ground plane at a height of approximately 20\,m. This setup allows us to obtain good coverage of the trees including the lower part of the canopy. The photogrammetric point clouds obtained in this way from the UAV are only included in the test set. Third, we use a RealSense d435i mounted on a mobile robot driving through the orchard rows to collect another part of our test set. Again, to obtain photogrammetric point clouds we use a bundle adjustment pipeline.
Furthermore, we label a few point clouds from the {Fuji-SfM} dataset~\cite{genemola2020compag,genemola2020db} to obtain a fourth test set from a different setup. This dataset is collected in~Agramunt in Catalunia, Spain with a Canon EOS 60D DSLR camera, followed by a bundle adjustment procedure to generate point clouds. In contrast to existing datasets in the agricultural domain~\cite{hani2020ral}~\cite{perezborrero2020cea}~\cite{weyler2024pami}, we specifically designed HOPS to have a remarkable domain shift between train and test sets, to more closely resemble real-world deployments. For this reason, we only included high-quality point clouds recorded with the TLS in the training set, while the test set includes a variety of sensors.

We report statistics about the different splits of our dataset in~\tabref{tab:data} and we show exemplary point clouds in~\figref{fig:data} to visualize the difference between training and test point clouds.
To summarize, our dataset consists of a training set and a validation set obtained with a TLS and four different test set obtained with different sensors (TLS, PhaseOne iXM-100, RealSense d435i, and EOS 60D) in different locations. 

For our dataset, we define 5 semantic classes, namely \textit{ground}, \textit{trunk}, \textit{canopy}, \textit{apple}, and \textit{pole}.  To formalize the panoptic segmentation task, we define the set of ``things'' classes (\textit{trunk}, \textit{apple}) and ``stuff'' classes (\textit{ground}, \textit{canopy}, \textit{pole}). Additionally, we define the \textit{tree} thing class which puts together apples and trunk belonging to the same tree.

To label the data, we split the aggregated point cloud of each orchard row in individual tiles with an average of 1~million points each. We manually label each tile in three stages using the online tool provided by \href{https://segments.ai/}{segments.ai}. 
One annotator labels each tile for semantic and two-level instance segmentation, ensuring consistency between instance levels. A second annotator then verifies label quality. Unlike image annotation, point cloud labeling requires frequent viewpoint changes, making it more time-consuming. The average effort consists of $4$ h per tile, totaling $525$ h for labeling and $175$ h for verification.



\section{Experimental Evaluation}
\label{sec:exp}

%
We present our experiments to show the capabilities of our method for hierarchical panoptic segmentation on real-world 3D point clouds. The results of our experiments also support our key claims, which are:
(i) our approach can jointly perform semantic, tree instance, and standard instance segmentation on real-world 3D data acquired with different sensors;
and
(ii) our skip connection scheme allows the CNN to exploit the underlying hierarchical relationship between the individual segmentation tasks, leading to better segmentation performance.

\subsection{Experimental Setup}
\textbf{Metrics}.
For semantic segmentation, we compute the intersection over union~(IoU)~\cite{everingham2010ijcv} over all five classes of our dataset, which we discussed in~\secref{sec:dataset}, and report the mean IoU~(mIoU) in the following tables. For the tree instance segmentation, we compute the single-class panoptic quality over the ``tree'' class, that we report as PQ$_\textrm{T}$. For the standard instance segmentation, we compute the panoptic quality over all classes, and report the mean panoptic quality as PQ in the tables. We also report the overall mean panoptic quality~(mPQ) as the average between PQ and PQ$_\textrm{T}$.

\begin{table}[t]
  \caption{Dataset statistics: number of samples per split, average number of points, fruits, and trunks per sample, and fruits per tree.}
  \centering
  \resizebox{0.99\linewidth}{!}{
  \begin{tabular}{ccccccc}
    \toprule
    \multirow{3}{*}{\textbf{}} & \textbf{Train} & \textbf{Val} &  \multicolumn{4}{c}{\textbf{Test}} \\ 
    \cmidrule(lr){2-2} \cmidrule{3-3} \cmidrule(lr){4-7}
     & TLS & TLS & UGV & UAV & SfM & TLS \\ 
     \midrule
    Samples            & 90    & 18    & 12    & 22    & 6     & 27 \\ 
    \arrayrulecolor{black!30}
    \midrule
    Points             & 0.8M  & 1M    & 0.6M  & 1.5M  & 1.8M  & 1M \\ 
    Fruits             & 90.2  & 99.2  & 71.8  & 99.1  & 232.3 & 39.9 \\ 
    Trunks             & 3.1   & 3.2   & 2.9   & 1.9   & 2.7   & 2.9 \\ 
    Fruits\slash tree  & 19.2  & 19.3  & 17.8  & 31.7  & 62.1  & 9.8 \\
    \arrayrulecolor{black}
    \bottomrule
  \end{tabular}
  }
  \label{tab:data}
  \vspace{-2em}
\end{table}

\begin{table*}[t]
  \caption{Results on the four different test sets of HOPS, and the average score across all sets and all tasks.}
  \centering

  \begin{tabular}{ccccccccccccccc}
    \toprule
    \multirow{3}{*}{\textbf{Approach}}           &  \multicolumn{3}{c}{\textbf{TLS}}  &  \multicolumn{3}{c}{\textbf{UAV}}  &  \multicolumn{3}{c}{\textbf{UGV}}  &  \multicolumn{3}{c}{\textbf{SfM}} & \multicolumn{2}{c}{\textbf{Average}} \\
    \cmidrule(lr){2-4} \cmidrule{5-7} \cmidrule(lr){8-10} \cmidrule(lr){11-13} \cmidrule(lr){14-15}
    & mIoU   & PQ    & PQ$_\textrm{T}$  & mIoU    & PQ    & PQ$_\textrm{T}$  & mIoU    & PQ    & PQ$_\textrm{T}$  & mIoU    & PQ    & PQ$_\textrm{T}$  & mIoU & mPQ  \\
    \midrule
    MaskPLS~\cite{marcuzzi2023ral} & 23.3 & 36.0 & - & 20.0 & 36.0 & - & 18.7 & 35.1 & - & 14.7 & 41.4 & - & \multirow{2}{*}{23.3} & \multirow{2}{*}{42.3}\\
    MaskPLS~\cite{marcuzzi2023ral} & - & - & 48.9 & - & - & 48.4 & - & - & \textbf{47.1} & - & - & \textbf{45.3} \\
    \arrayrulecolor{black!30}
    \midrule
    ForestPS~\cite{malladi2025icra} & 52.0 & 49.4 & - & 64.3 & \textbf{62.8} & - & 40.2 & \textbf{37.7} & - & 47.0 & \textbf{46.2} & - & \multirow{2}{*}{50.9} & \multirow{2}{*}{27.5}\\
    ForestPS~\cite{malladi2025icra} & - & - & 21.5 & - & - & 2.1 & - & - & 0 & - & - & 0  \\
    \arrayrulecolor{black}
    \midrule
    Ours & \textbf{57.7} & \textbf{49.5} & \textbf{55.4} & \textbf{65.2} & 51.9 & \textbf{48.9} & \textbf{42.6} & 35.2 & 42.1 & \textbf{47.7} & 34.5 & 39.1 & \textbf{53.3} & \textbf{44.6} \\
    \bottomrule
  \end{tabular}
  \label{tab:test_set_res}
  \vspace{-1em}
\end{table*}

\begin{table}[t]
  \caption{Results on the validation set. We report the average iteration time (s/it) on a GPU NVIDIA RTX A5000.}
  \centering
  \begin{tabular}{ccccccc}
    \toprule
    \multirow{3}{*}{\textbf{Approach}} &  \multicolumn{4}{c}{\textbf{TLS}} & \multirow{3}{*}{Param} & \multirow{3}{*}{s/it} \\
    \cmidrule(lr){2-5}
     & mIoU    & PQ    & PQ$_\textrm{T}$ & mPQ\\
    \midrule
    MaskPLS~\cite{marcuzzi2023ral} & 28.2 & 42.8 & - & \multirow{2}{*}{45.6} & \multirow{2}{*}{1.8M} & \multirow{2}{*}{4.0}\\
    MaskPLS~\cite{marcuzzi2023ral} & - & - & 48.9 \\
    \arrayrulecolor{black!30}
    \midrule
    ForestPS~\cite{malladi2025icra} & 58.9 & \textbf{62.2} & - & \multirow{2}{*}{32.3} & \multirow{2}{*}{1.5M} & \multirow{2}{*}{1.8}\\
    ForestPS~\cite{malladi2025icra} & - & - & 2.3 \\
    \arrayrulecolor{black}
    \midrule
    Ours & \textbf{70.9} & 60.8 & \textbf{52.1} & \textbf{56.5} & 19.4M & 2.0\\
    \bottomrule
  \end{tabular}
  \label{tab:val_set_res}
  \vspace{-1.5em}
\end{table}

\textbf{Training details and parameters}.
In all experiments, we use AdamW~\cite{loshchilov2017arxiv} with weight decay of $0.99$ with an initial learning rate of $5 \cdot 10^{-3}$. We set the weights of the loss function to $w_1 = w_2 = w_3 = 1$. We train our approach for 500 epochs. We set batch size of 1. We use voxel downsampling to $3 \, \mathrm{mm}$. Due to practical constraints, the training set is limited to TLS-acquired data. However, to mitigate potential overfitting to TLS-specific characteristics and improve generalization to non-TLS domains, we apply extensive data augmentation techniques including scale, rotation around all axes, shear, color jittering, and elastic deformation~\cite{castro2018bhi}. We tuned all hyperparameters on the validation set.

All baselines can only address one instance segmentation task at a time and, thus, need to be trained twice to solve both instance-level tasks. Our method addresses both instance segmentation tasks with a single training run.

\subsection{Experiments on Hierarchical Panoptic Segmentation}
While our PQ is inferior to ForestPS, our approach also yields good performance on tree instance segmentation, on which the dedicated training of ForestPS fails completely, achieving results below $3 \%$ on four out of five splits.
The first experiment evaluates the performance of our approach and its outcomes support the claim that our approach can jointly perform semantic, tree instance, and standard instance segmentation on real-world 3D data. For this experiment, we compare our approach to existing baselines for panoptic segmentation: MaskPLS~\cite{marcuzzi2023ral} is a transformer-based approach that extends Mask2Former~\cite{cheng2021cvpr} to 3D point clouds, while ForestPS~\cite{malladi2025icra} is a convolutional neural network approach based on MinkUNet~\cite{choy2019cvpr}, originally designed for tree segmentation in a forest environment. The baselines are trained for only one of the two instance segmentation tasks at a time, as they do not support hierarchical multi-level instance segmentation. As shown in ~\tabref{tab:test_set_res} for the test set, and~\tabref{tab:val_set_res} for the validation set, our approach outperforms the baselines on both splits in terms of mIoU and on most PQ$_\textrm{T}$, while ForestPS achieves the best PQ on most splits. However, our approach performs better than the baselines on mPQ.

\begin{figure*}
  \centering
  \includegraphics[width=.99\linewidth]{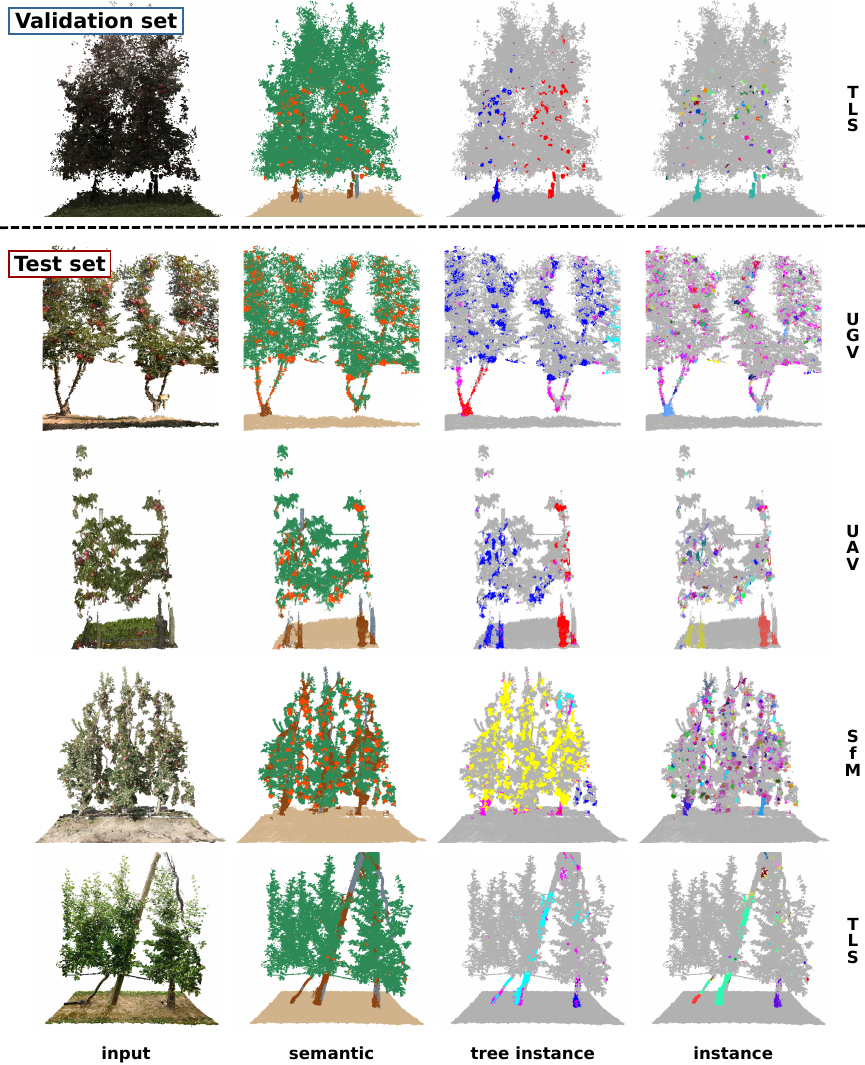}
  \caption{Qualitative results of our approach for hierarchical panoptic segmentation. The first row shows validation set results, the other rows show test set result. We report the input point cloud with all three segmentation predictions, and one result for each sensor. In the semantic prediction, different colors indicate different categories. In the instance predictions, different colors indicate different instances.}
  \label{fig:quali}
\end{figure*}

We show qualitative results in~\figref{fig:quali}. As explained in~\secref{sec:dataset}, only the validation set is recorded in the same time period and with the same sensor as the training set. The test sets, in contrast, are recorded either 2 years later also using a TLS, or with different sensors (UGV and UAV), or belong to entirely different datasets (SfM). This motivates the performance gap between the validation and the test set. We believe that this aspect pushes researchers to build models that yield good generalization performance and can be used in different conditions. In~\tabref{tab:test_set_res}, we also report average scores for mIoU and PQ across all four splits and both instance segmentation tasks, to show how different approaches deal with both tasks at the same time.

\subsection{Ablation Studies} \label{sec:ablations}
In the ablation study, we aim to validate our claim that the proposed skip connection scheme allows the CNN to exploit the underlying hierarchical relationship between the segmentation tasks, leading to better segmentation performance. We perform this experiment on the validation set. We compared to other skip connection schemes, and report results in~\tabref{tab:ablation}. The simplest approach does not use skip connections (denoted as A in~\tabref{tab:ablation}), which leads to poor performance. This is expected, since the decoders do not get any high-level information from features coming from earlier stages, which harms gradient flow. The second approach uses skip connections only from decoder to decoder, excluding the contributions from the encoder (denoted as B in~\tabref{tab:ablation}). This approach, despite exploiting the hierarchy among tasks with decoder-based skip connections, does not yield good performance. The third approach is the standard UNet-like skip connections from the encoder to all decoders (denoted as C in~\tabref{tab:ablation}). In this case, all decoders obtain the same information from the encoder. This proves more effective than the decoder-only skip connections. This is probably due to the fact that restoring high-level information is important, and using decoder-only skip connections does not help gradient propagation. Thus, our method using skip connections from the encoder and from decoder to decoder (denoted as D in~\tabref{tab:ablation}) performs better than all others, as on one hand it benefits from the high-level features coming from the encoder, and on the other hand it exploits the underlying hierarchy among segmentation tasks. Interestingly, it also provides better results on semantic segmentation, despite there is no difference in the skip connection scheme for the semantic decoder, as the hierarchy affects only the instance decoders. This suggests that having information coming from the instance decoders indirectly affecting the encoder and the semantic decoder via backpropagation is useful for semantic segmentation performance.

\begin{table}[t]
  \caption{Ablation study on the validation set. Comparison between different skip connections schemes.}
  \centering

  \begin{tabular}{cccccc}
    \toprule
    & \textbf{Skip Connections}            & \textbf{mIoU}    & \textbf{PQ}    & \textbf{PQ$_\textrm{T}$} & \textbf{mPQ}        \\
    \midrule
    A & None & 55.8 & 42.8 & 47.7 & 45.3\\    
    B & Decoder & 64.4 & 53.4 & 51.3 & 52.4 \\    
    C & Encoder & 69.8 & 58.8 & 50.9 & 54.9 \\    
    \midrule
    D & Encoder+Decoder & \textbf{70.9} & \textbf{60.8} & \textbf{52.1} & \textbf{56.5} \\    
    \bottomrule
  \end{tabular}
  \label{tab:ablation}
  \vspace{-1em}
\end{table}
\section{Conclusion}
\label{sec:conclusion}

In this paper, we presented a novel approach for hierarchical 3D panoptic segmentation on point cloud data. By means of a novel skip connections scheme, our method exploits the underlying hierarchy among different segmentation tasks, and is able to yield semantic segmentation, tree instance segmentation, where a tree is defined as a trunk and all the apples belonging to it, and standard instance segmentation at the same time. Thanks to the proposed skip connection scheme in our architecture, our approach achieves state-of-the-art results, surpassing or being comparable to existing task-specific baselines, despite that they can deal with only one instance segmentation task at a time. Additionally, we introduce a novel point cloud dataset of real apple orchards, called HOPS, labeled for hierarchical panoptic segmentation. Our dataset includes data recorded with different sensors over the course of two years. In our experimental evaluation, we supported all claims made in this paper and showed that our proposed dataset is challenging.

\addtolength{\textheight}{-.5cm} 


\bibliographystyle{plain_abbrv}

\bibliography{glorified,new}

\end{document}